%% file: example_paper.tex
\documentclass{article}

\usepackage{microtype}
\usepackage{graphicx}
\usepackage{subcaption}
\usepackage{booktabs} 

\usepackage{hyperref}

\usepackage[accepted]{icml2026}

\usepackage{amsmath}
\usepackage{amssymb}
\usepackage{mathtools}
\usepackage{amsthm}

\usepackage[capitalize,noabbrev]{cleveref}

\theoremstyle{plain}
\newtheorem{theorem}{Theorem}[section]
\newtheorem{proposition}[theorem]{Proposition}
\newtheorem{lemma}[theorem]{Lemma}
\newtheorem{corollary}[theorem]{Corollary}
\theoremstyle{definition}
\newtheorem{definition}[theorem]{Definition}
\newtheorem{assumption}[theorem]{Assumption}
\theoremstyle{remark}
\newtheorem{remark}[theorem]{Remark}

\usepackage[textsize=tiny]{todonotes}

\usepackage{times}
\usepackage{latexsym}
\usepackage{pifont}
\input{math_commands.tex}

\usepackage{ulem}
\usepackage{hyperref}
\usepackage{url}
\usepackage{tabularx}
\usepackage{graphicx}
\usepackage{colortbl}
\usepackage{comment}
\usepackage{multirow}
\usepackage[table,xcdraw]{xcolor}
\usepackage{wrapfig}
\usepackage{multicol}
\usepackage{amsfonts}
\usepackage{dsfont} 
\usepackage{xurl}
\usepackage{soul} 

\usepackage{amsmath,amssymb}
\usepackage{tcolorbox}

\usepackage{caption}
\tcbset{sharp corners, colback=white, boxrule=0pt}
\tcbuselibrary{skins,breakable}
\newtcolorbox{transcriptbox}{
enhanced, breakable, blanker,
width=\linewidth,
left=0pt, right=0pt, top=6pt, bottom=6pt,
borderline north={0.6pt}{0pt}{black},
borderline south={0.6pt}{0pt}{black}
}

\usepackage[T1]{fontenc}

\usepackage[utf8]{inputenc}

\usepackage{microtype}

\usepackage{inconsolata}

\usepackage{graphicx}

\icmltitlerunning{Extending RLVR to Open-Ended Tasks via Verifiable Multiple-Choice Reformulation}

\begin{document}

\twocolumn[
  \icmltitle{Extending RLVR to Open-Ended Tasks via Verifiable Multiple-Choice Reformulation}
%


  \icmlsetsymbol{equal}{*}

  \begin{icmlauthorlist}
    \icmlauthor{Mengyu Zhang}{equal,comp}
    \icmlauthor{Siyu Ding}{equal,comp}
    \icmlauthor{Weichong Yin}{comp}
    \icmlauthor{Yu Sun}{comp}
    \icmlauthor{Hua Wu}{comp}
  \end{icmlauthorlist}

  \icmlaffiliation{comp}{Baidu Ernie Team, Beijing, China}


  \icmlcorrespondingauthor{Siyu Ding}{dingsiyu@baidu.com}

  \icmlkeywords{Machine Learning, ICML}

  \vskip 0.3in
]



\printAffiliationsAndNotice{\icmlEqualContribution}

\begin{abstract}

Reinforcement Learning with Verifiable Rewards (RLVR) has demonstrated great potential in enhancing the reasoning capabilities of large language models (LLMs). However, its success has thus far been largely confined to the mathematical and programming domains with clear and automatically checkable outcomes. Reinforcement learning on open-ended tasks (e.g., creative writing and subjective Q\&A) continues to rely on reward models due to the absence of verifiable solutions. This raises a key question: \textit{how can we extend RLVR to strengthen reasoning in open-ended tasks regardless of the absence of the unambiguous ground truth?}
To overcome this challenge, we 
introduce \textbf{V}erifiable \textbf{M}ultiple-Choice \textbf{R}eformulation for Reinforcement Learning from Verifiable Rewards (\textbf{VMR-RLVR}),
a novel training strategy that restructures open-ended data into verifiable multiple-choice formats, enabling effective training even in the absence of explicit ground truth. 
Experimental results on multiple benchmarks validate the effectiveness of our method in improving LLM performance on open-ended tasks.
Notably, across seven open-ended benchmarks, our VMR-RLVR training delivers an average gain of 3.29 points over the RL with reward model. Code will be released upon acceptance to facilitate reproducibility.
\end{abstract}

\section{Introduction}
Reinforcement Learning with Verifiable Rewards (RLVR) has recently emerged as a powerful paradigm for enhancing the reasoning capabilities of Large Language Models (LLMs) \cite{DBLP:journals/corr/abs-2412-16720,DBLP:journals/corr/abs-2501-12948,DBLP:journals/corr/abs-2503-24290,DBLP:journals/corr/abs-2501-12599,DBLP:journals/corr/abs-2410-15115,DBLP:journals/corr/abs-2411-15124,DBLP:journals/corr/abs-2504-20571}. In RLVR, a Large Language Model (LLM) serves as a policy agent, generating a Chain of Thought (CoT) as a structured sequence of actions while receiving real-time feedback on answer correctness from deterministic verifiers. This paradigm not only shows the potential of scaling test-time computation for tackling complex problems \cite{DBLP:journals/corr/abs-2504-20571}, but also holds the promise of endowing LLMs with the ability to learn from experience through free exploration, potentially leading to general artificial intelligence \cite{DBLP:journals/corr/abs-2506-18254}.
\begin{figure}[t]
    \centering
    \includegraphics[width=1.0\linewidth]{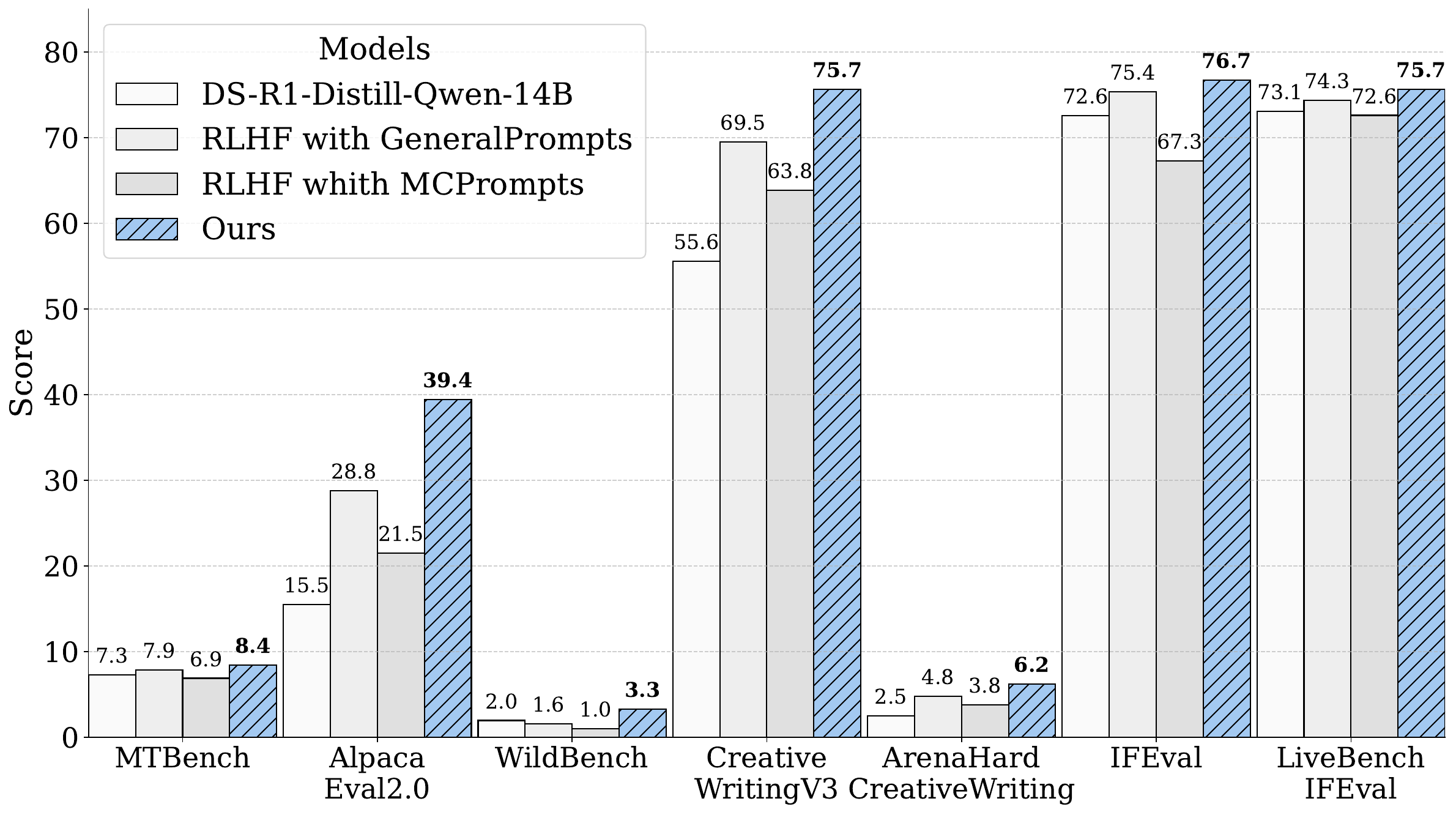}
    \vspace{-3pt}
    \caption{Overall performance on seven open-ended benchmarks. Our proposed VMR-RLVR method consistently enhances performance across multiple benchmarks and outperforms strong RLHF baselines. }
    \label{fig:bar_performance}
    \vspace{-10pt}
\end{figure}

\begin{figure*}[t]
    \vspace{-3pt}
    \centering
    \includegraphics[width=0.98\linewidth]
    {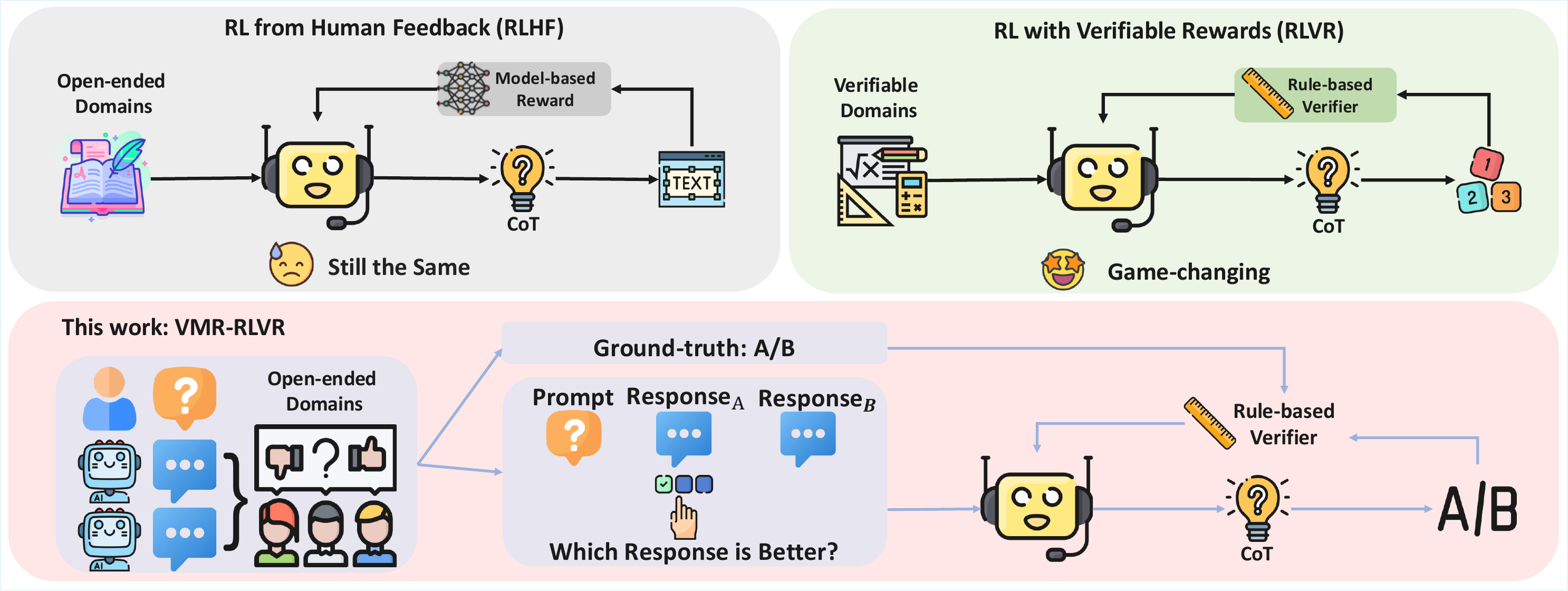}
    \caption{Overview of the VMR-RLVR framework. 
    For each open-ended \textit{prompt}, we construct a candidate set consisting of a \textit{chosen response} and a \textit{rejected response}. The two options are randomly ordered to form a multiple-choice question, and the LLM is required to identify the correct one. A verifier then provides binary feedback, enabling RLVR-style optimization in open-ended domains without explicit ground-truth references.
}
\label{fig:method}
\vspace{-10pt}
\end{figure*}

At its core, RLVR rests on using data that are challenging for a model to solve but can be easily and objectively verified. Representative domains include mathematics and competitive programming, where solutions can be validated automatically: mathematical answers by matching to the canonical solution, and code solutions by executing against a suite of test cases on an online sandbox environment \cite{DBLP:journals/corr/abs-2505-21493,DBLP:journals/corr/abs-2506-18254}. While the RLVR algorithm has achieved remarkable success, its effectiveness is constrained by reliance on prompts with precise reward signals provided by domain-specific verifiers. Recent works attempt to extend the RLVR paradigm to more general domains (such as science) and yield measurable gains \cite{DBLP:journals/corr/abs-2506-18254}. However, these studies systematically overlook open-ended tasks (e.g., creative writing and subjective Q\&A), where no unambiguous ground truth exists. OpenAI shows the composition of user messages over time in \cite{chatterji2025people}, in which the Writing topics have remained constant at roughly 25\% of overall usage. More than that, there are about 40\% of all work-related messages are Writing when restricting the sample to only work-related messages. Therefore, strengthening reasoning in open-ended tasks is of paramount importance, thereby broadening the applicability of RLVR to a much wider range of real-world scenarios.

Extending RLVR to open-ended domain tasks in an unvarnished way is impractical, given the  high diversity and complexity of natural language typically preclude an unambiguous ground truth for such tasks.
Current researches \cite{DBLP:journals/corr/abs-2505-14652,lu2025writing,gunjal2025rubrics,DBLP:journals/corr/abs-2507-18624,DBLP:journals/corr/abs-2507-17746,DBLP:journals/corr/abs-2509-20357} attempt to address this problem by training specialized LLMs as verifier models. Notwithstanding their widespread adoption, reward models suffer from three principal limitations: 
(1) training LLMs to function as general reward evaluators requires non-trivial and substantial data annotation, and the resulting reward signals are frequently of limited quality in practice; 
(2) they are vulnerable to reward hacking, with a tendency to overfit superficial artifacts and spurious correlations; 
(3) the integration of a separate verifier model introduces additional architectural complexity and increases computational cost in the RL training regime.
Consequently, the scalability limitations of existing RLVR methods prevent leveraging rich open-ended domain data and curtail the potential of broader reasoning capabilities.
This motivates a critical research question: \textit{how can we extend RLVR to strengthen reasoning in open-ended tasks regardless of the absence of the unambiguous ground truth?}

To tackle this challenge, we introduce \textbf{VMR-RLVR} (Verifiable Multiple-Choice Reformulation for Reinforcement Learning from Verifiable Rewards), a framework that extends RLVR to open-ended domain tasks without external verifiers. The responses to open-ended-domain prompts cannot be verified, given the absence of unambiguous ground truth. That is, in a \textit{(prompt $\Rightarrow$ response)} pair, 
the \textit{response} is unverifiable. A straightforward approach is to express the pair in a verifiable format by task type conversion. Among numerous task types, multiple-choice and true/false tasks stand out for their inherent verifiability, which are not limited by the task domain and have broad applicability. Therefore, we convert the open-ended data pair \textit{(prompt $\Rightarrow$ response)} to \textit{((prompt, response$_{1}$, response$_{2}$, ..., response$_{N}$)~$\Rightarrow$ which one is best?)}, the detail is depicted in Figure~\ref{fig:method}. This transformation enables the application of RLVR-style optimization even when the unambiguous ground truth does not exist. 
Notably, the RLHF objective trained on a small amount of prompts is incorporated into VRM-RLVR optimization with a modest weighting coefficient to alleviate the collapse of response lengths observed when training exclusively on multiple-choice data.
VMR-RLVR preserves the core strengths of RLVR, such as clear reward signals and reasoning-oriented training, and further extends its applicability to open-ended tasks without domain-specific verifiers. 

\begin{figure}[t]
    \vspace{-3pt}
    \centering
    \includegraphics[width=1.0\linewidth]
    {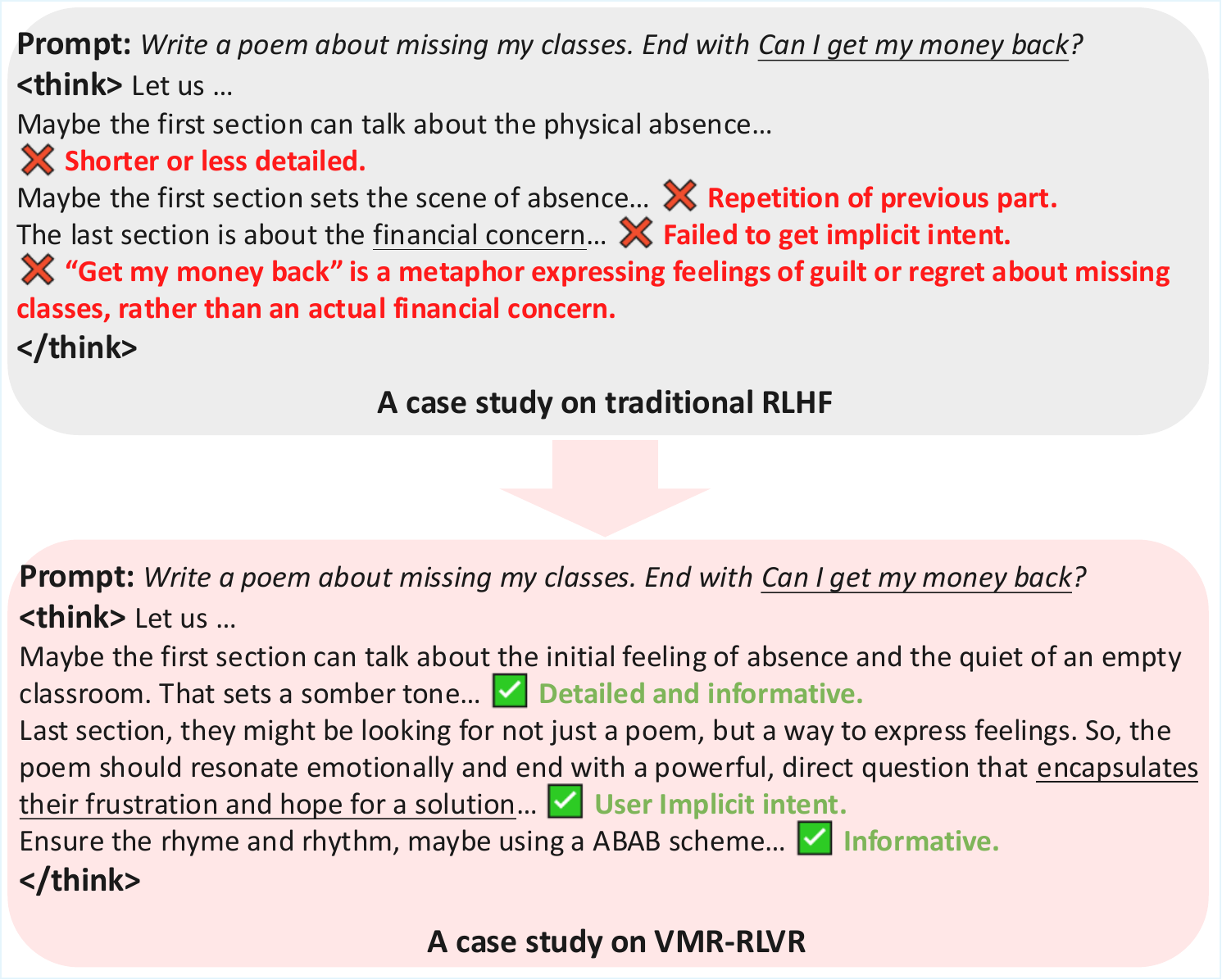}
    \caption{ A case study shows that VMR-RLVR improves open-ended reasoning, producing more informative thinking processes with less repetition and stronger alignment to users’ implicit intent.
}
\label{fig:CaseStudy}
\vspace{-10pt}
\end{figure}

We conduct extensive experiments across multiple open-ended benchmarks to evaluate the effectiveness of our approach. Results demonstrate that our method not only improves task performance but also substantially enhances the reasoning trajectories produced by LLMs. Concretely, leveraging DeepSeek-R1-Distill-Llama-8B and DeepSeek-R1-Distill-Qwen-14B as the base model, our method achieves improvements of 2.6 and 3.29 average points over the RLHF across seven open-ended benchmarks respectively. These findings provide compelling evidence that RLVR can be extended to open-ended domain tasks without an unambiguous ground truth, by appropriate task reformulation. Figure \ref{fig:CaseStudy} demonstrates the improvement in the reasoning capabilities of VRM-RLVR for open-ended-domain data. More broadly, our work suggests that the frontier of RLVR need not be restricted to mathematical or programming domains, but can be extended to a much wider range of real-world scenarios. 

The contributions of this paper are threefold:
\begin{itemize}
    \item We highlight the underexplored issue of whether RLVR can be extended to strengthen reasoning in open-ended tasks regardless of the absence of the unambiguous ground truth. To the best of our knowledge, this is the first attempt to expand RLVR into open-ended domains without exploring advanced reward signals.
    \item We introduce VMR-RLVR, a simple yet effective framework that restructures open-ended domain data into verifiable multiple-choice formats, enabling RLVR to operate beyond strictly verifiable STEM domains. 
    \item We empirically validate the effectiveness of our method. The experimental results indicate that our method significantly enhances reasoning capabilities and results in an average improvement of 3.29 points across seven different benchmarks. 
\end{itemize}

\section{Method}
\label{sec:method}


\subsection{Preliminaries: RLHF and RLVR}
\textbf{RLHF} aligns a language model’s outputs with human preferences by training a reward model on human judgments. A language model parameterized by $\pi_\theta$ defines a conditional auto-regressive generator that produces a response $y$ conditioned on an input $x \in \mathcal{D}$. A defined reward function $\mathcal{R}(y|x)$ assigns a bounded score as a reward signal to the ($x$, $y$) pair. The goal is to train a policy $\pi_\theta$ that generates a response $y$ maximizing the RL objective (we omit the KL divergence for brevity):
%
\begin{equation}
    \mathcal{J}_{\text{RLHF}}(\theta) = \mathbb{E}_{x \in \mathcal{D}, ~y \sim \pi_\theta(\cdot|x)} \left[ \mathcal{R}(y|x) \right],
    \label{eq:rl_objective}
\end{equation}


\textbf{RLVR} is a general reinforcement learning paradigm in which a \textit{rule-based verifier} assigns a scalar score as a reward signal to each generated response. Specifically, given an input $x \in \mathcal{D}$, the generation $o$ produced by policy $\pi_\theta$ is typically decomposed into a reasoning trajectory $z$ and a final answer $y$ (i.e., $o$=($z$,$y$)). Then, the verifier $\mathcal{F}_{\text{verifier}}$ is required to check whether the generated answer $y$ passes the test defined by the ground-truth answer $y^\star$ and assigns a binary reward. The training objective can be written as follows:
\begin{equation}
    \mathcal{J}_{\text{RLVR}}(\theta) = \mathbb{E}_{x \in \mathcal{D},~z,y \sim \pi_\theta(\cdot|x)} \left[ \mathcal{F}_{\text{verifier}}(y , y^\star) \right]
\label{eq:verifier_objective}
\end{equation}

\begin{figure}[t]
    \centering
    \includegraphics[width=0.9\linewidth]{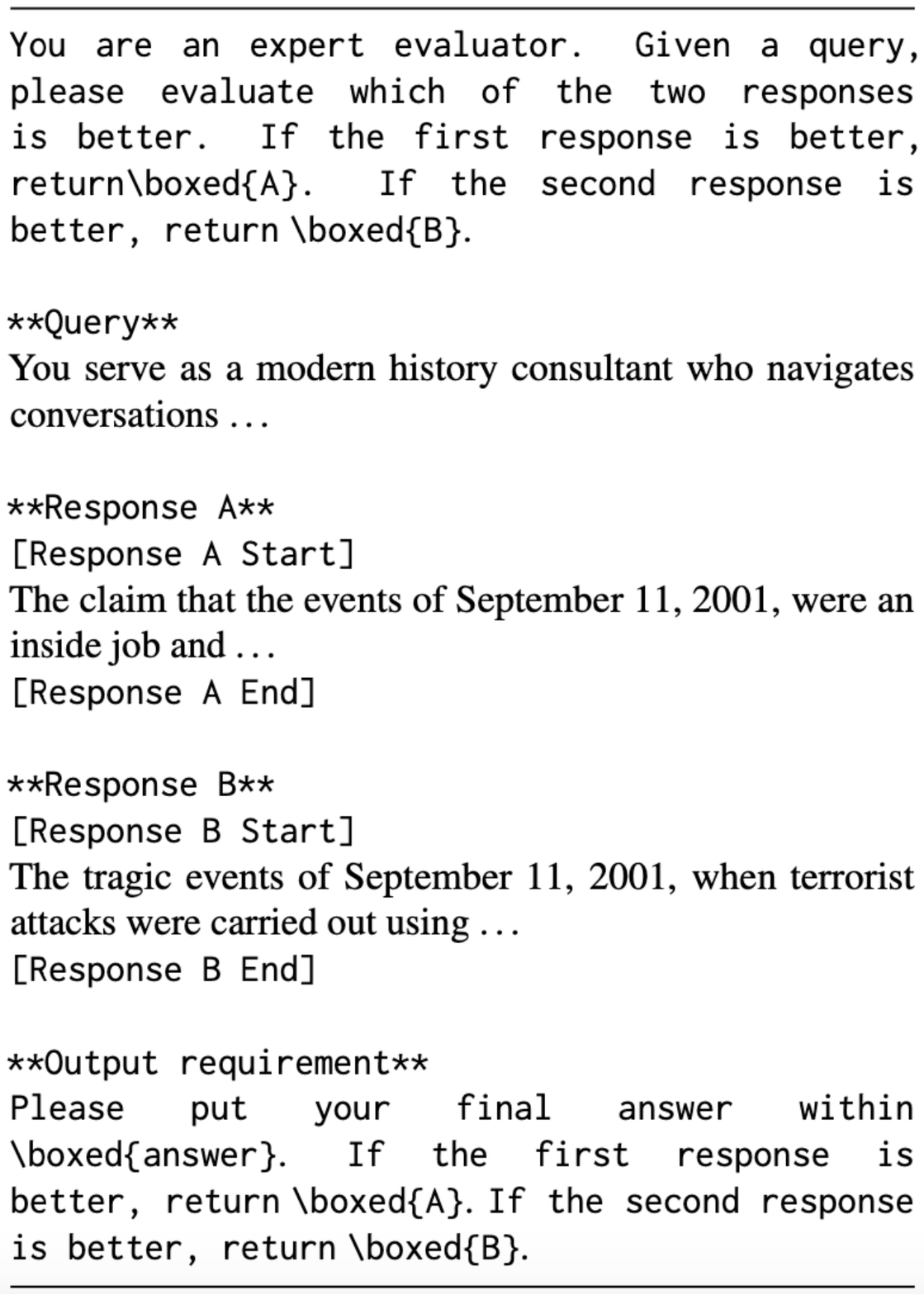}
    \caption{ Verifiable Multiple-Choice Reformulation (VMR) template.}
    \label{fig:vmr}
\end{figure}

\subsection{Problem Formulation: Restructuring Open-Ended Task Data into Verifiable Formats}



The RLVR framework has achieved remarkable success in mathematics and competitive programming domains; its effectiveness is constrained by reliance on prompts with precise reward signals provided by domain-specific verifiers. In open-ended domain tasks, high-quality responses are inherently diverse and there is no unambiguous ground truth against which correctness can be effectively checked. This lack of verifiable verifiers poses a fundamental challenge: \textit{how can we extend RLVR to strengthen reasoning in open-ended domain tasks regardless of the absence of the unambiguous ground truth?}

To address this challenge, we reformulate open-ended task data into \textit{multiple-choice verification problems}. Specifically, given an input prompt $x \in \mathcal{D}$ and its corresponding open-ended generation space $\mathcal{S}$, we construct a candidate response set $\mathfrak{S} = \{s_{1}, s_{2}, ..., s_{N}\}$ with cardinality $N$. The goal is to identify the optimal response from among the candidate responses set. To avoid positional bias, the order of the candidate responses set $\mathfrak{S} = \{s_{1}, s_{2}, ..., s_{N}\}$ is randomized when forming the multiple-choice question. For example, when $N$ is set to 2, the candidate responses set $\mathfrak{S} = \{s^{+}, s^{-}\}$ consists of one chosen response $s^{+}$ and one rejected response $s^{-}$. When forming the multiple-choice question: if $s^{+}$ is placed first, the correct option corresponds to ``A''; otherwise, it corresponds to ``B''. The policy model is then required to select the better one between the two options (Figure \ref{fig:vmr} presents an instruction), and then a verifier needs to check whether the final answer $y$ passes the test defined by the ground-truth answer $y^{\star}$. The resulting verifier is defined as:
\begin{equation}
\mathcal{F}_{\text{verifier}}(y, y^{\star}) =
\begin{cases}
1, & \text{if selected option denotes } s^{+}, \\
0, & \text{if selected option denotes } s^{-}.
\end{cases}
\label{eq:choice_reward}
\end{equation}

\begin{figure}[t]
    \centering
    \includegraphics[width=1.0\linewidth]{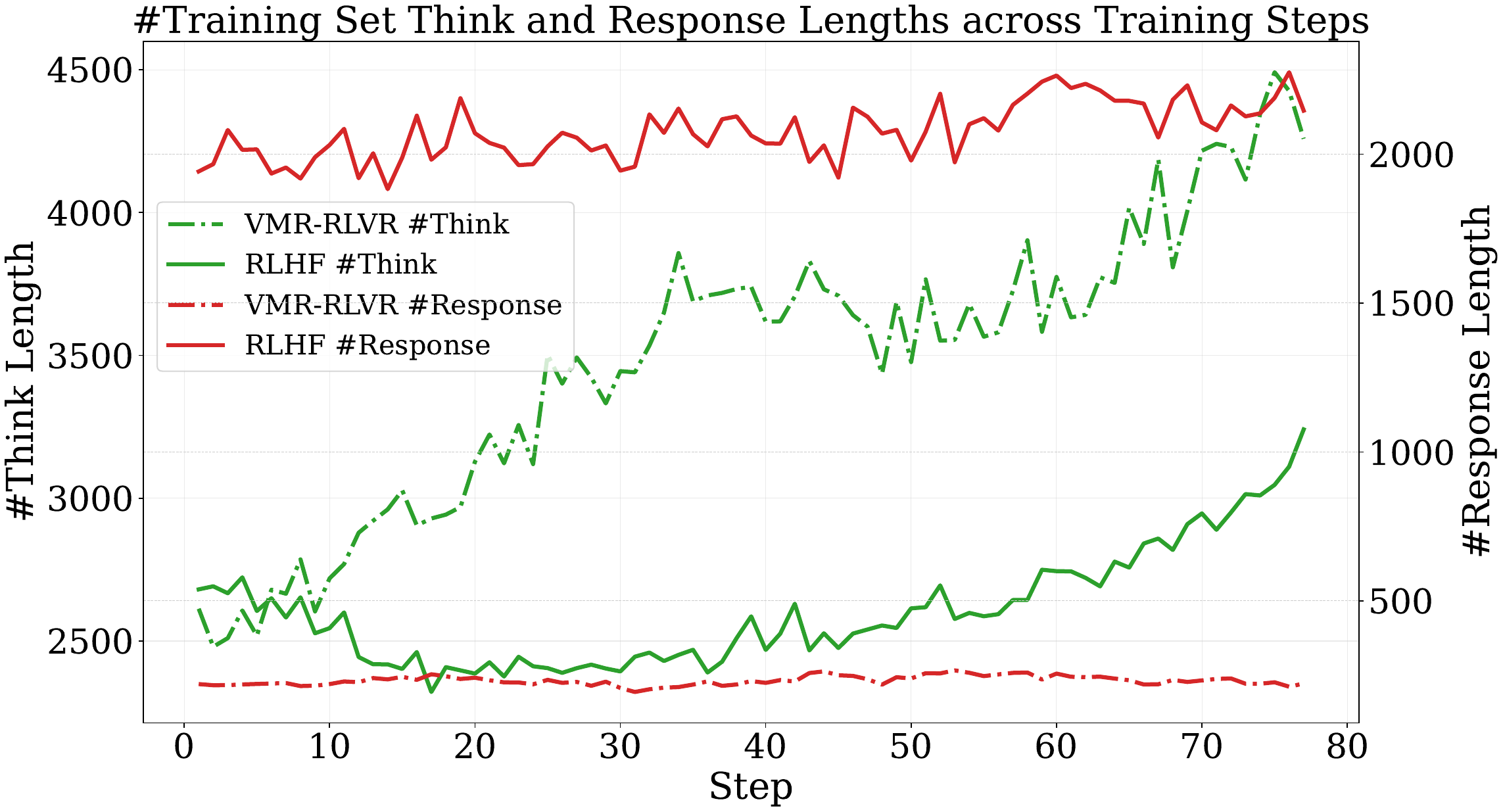}
    \caption{ Comparison of think length and response length in the training sets of preliminary VMR-RLVR and RLHF models. 
    In this context, preliminary VMR-RLVR refers to RLVR with multiple-choice questions, and RLHF denotes the use of RLHF with the GeneralPrompts dataset. 
    The solid line represents RLHF, while the dashed line indicates preliminary VMR-RLVR. 
    The green line signifies the average think length, and the red line denotes the average response length.
    }
    \vspace{5pt}
    \label{fig:pure_rlvr_think_response}
\end{figure}

Throughout our preliminary experiments, we have observed that there is a substantial improvement of reasoning capabilities on open-ended domain tasks when training the policy model $\pi_{\theta}$ exclusively on multiple-choice prompts. However, an unanticipated problem is that the collapse of response lengths significantly restrict the response pattern. As illustrated in Figure~\ref{fig:pure_rlvr_think_response}, 
the response length in preliminary VMR-RLVR experiments  (red dashed line) remains consistently short, substantially shorter than that in RLHF experiments (red solid line). Consequently, we incorporated an RLHF objective, derived from a limited general prompt set, into the VRM-RLVR optimization, with a small weighting coefficient to mitigate this issue. Concretely, given an input prompt $x$ and the candidate responses set $\mathfrak{S} = \{s_{1}, s_{2}, ..., s_{N}\}$, the RL
objective is defined as:
\begin{equation}
    \begin{aligned}
    \mathcal{J}_{\text{VMR-RLVR}}(\theta) &= \mathbb{E}_{x \in \mathcal{D},~z,y \sim \pi_\theta(\cdot|(x, \mathfrak{S}))} \left[ \mathcal{F}_{\text{verifier}}(y , y^\star) \right] \text{+}\\
    & ~~~~~\alpha \cdot \mathbb{E}_{x \in \mathcal{D}_{s}, ~z,y \sim \pi_\theta(\cdot|x)} \left[ \mathcal{R}(y|x) \right],
    \end{aligned}
\label{eq:verifier_objective}
\end{equation}
where $\mathcal{D}_{s}$ is a subset of $\mathcal{D}$, and $\alpha$ is the weighting coefficient for RLHF objective.
VMR-RLVR innovatively establish the verifiability on open-ended domain tasks, where correctness is well-defined within the multiple-choice questions, even if there is no unambiguous ground truth exist for original prompts.


\subsection{Comparison to Existing Approaches}

Comprehensive comparisons between the proposed
VMR-RLVR and existing approaches are presented in Table \ref{tab:compare-method}. 

\begin{table}[]
\caption{ Comprehensive comparisons between the proposed method and existing approaches are presented. }

\resizebox{1.0\linewidth}{!}{

\begin{tabular}{lccc}
\hline
\rowcolor[HTML]{FFFFFF} 
\multicolumn{1}{c}{\cellcolor[HTML]{FFFFFF}{\color[HTML]{1C1D1F} Methods}}                           & \textbf{Open-ended}        & \textbf{Verifier}          & \textbf{Rely on ground-truth} \\ \hline
\rowcolor[HTML]{FFFFFF} 
VeriFree\cite{DBLP:journals/corr/abs-2505-21493}                                    & \textcolor{red!50!black}{\ding{56}} & \textcolor{green!50!black}{\ding{52}} & \textcolor{green!50!black}{\ding{52}}    \\
\rowcolor[HTML]{EFEFEF} 
RLPR \cite{DBLP:journals/corr/abs-2506-18254}                                       & \textcolor{red!50!black}{\ding{56}} & \textcolor{green!50!black}{\ding{52}} & \textcolor{green!50!black}{\ding{52}}    \\
\rowcolor[HTML]{FFFFFF} 
{\color[HTML]{333333} Rubric \cite{DBLP:journals/corr/abs-2507-17746}}  & \textcolor{green!50!black}{\ding{52}} & \textcolor{red!50!black}{\ding{56}} & \textcolor{red!50!black}{\ding{56}}    \\
\rowcolor[HTML]{EFEFEF} 
{\color[HTML]{333333} Checklist \cite{DBLP:journals/corr/abs-2507-18624}} & \textcolor{green!50!black}{\ding{52}} & \textcolor{red!50!black}{\ding{56}} & \textcolor{red!50!black}{\ding{56}}    \\
\rowcolor[HTML]{FFFFFF} 
RLMT \cite{DBLP:journals/corr/abs-2509-20357}    & \textcolor{green!50!black}{\ding{52}} & \textcolor{red!50!black}{\ding{56}} & \textcolor{red!50!black}{\ding{56}}    \\
\rowcolor[HTML]{FFDFDD} 
Ours                                                                                                 & \textcolor{green!50!black}{\ding{52}} & \textcolor{green!50!black}{\ding{52}} & \textcolor{green!50!black}{\ding{52}}    \\ \hline
\end{tabular}
}
\vspace{-5pt}
\label{tab:compare-method}
\end{table}

In open-ended settings, most approaches use learned reward models to provide reward feedback, including rubric-based rewards \cite{DBLP:journals/corr/abs-2507-17746}, checklist-based rewards \cite{DBLP:journals/corr/abs-2507-18624}, and RLMT-based methods \cite{DBLP:journals/corr/abs-2509-20357}. These methods rely on trained reward models to generate bounded scores rather than on rule-based verifiers. 

In RLVR settings, both our method and recent works such as \textsc{VeriFree} \cite{DBLP:journals/corr/abs-2505-21493} and \textsc{RLPR} \cite{DBLP:journals/corr/abs-2506-18254} aim to overcome the fundamental limitation 
of RLVR—its reliance on explicit verifiers. However, \textsc{VeriFree} and \textsc{RLPR} replace the verifier with the model’s own conditional probability of the ground truth $y^\star$. They still fundamentally rely on the existence of unique ground-truth solutions, which makes them unsuitable for open-ended tasks such as creative writing and subjective Q\&A. In contrast, our method addresses the verifier limitation from a different perspective. Instead of relying on unambiguous ground truth, we restructure open-ended tasks into multiple-choice questions.

\section{Experiments}

%
%
\subsection{Experimental Setup}

\textbf{Models.}  
We use two reasoning models, DeepSeek-R1-Distill-Qwen-14B and DeepSeek-R1-Distill-Llama-8B \cite{DBLP:journals/corr/abs-2501-12948}, as base models to evaluate whether VMR-RLVR improves the reasoning capabilities for open-ended domain tasks.
Additionally, we conduct experiments with a non-reasoning model, Qwen2.5-14B-Instruct \cite{qwen2.5}, to explore how VMR-RLVR performs on models without built-in reasoning capabilities. 
All experiments are conducted under the GRPO framework \cite{DBLP:journals/corr/abs-2402-03300}.
To ensure the validity of the experiments, we controlled for sample size and computational resources by using the identical quantities of training samples and the same computational budget across all experiments.


\textbf{Training Data and Rewarding Methods.} 
We build several datasets to evaluate our approach. 
To economize on resources, we create MCQuestions by converting existing preference data into verifiable multiple-choice questions.
\begin{itemize}

    \item \textbf{MCQuestions Dataset}  
    contains approximately 20k triples drawn from preference datasets, including \textit{Magpie Pro Standard} \cite{DBLP:conf/iclr/XuJNDP0L25}, \textit{RM\_OA\_HH} \cite{pvduy_rm_oa_hh} and \textit{Multifaceted CollectionRM} \cite{DBLP:conf/nips/LeePKS24}.    
    Each sample is converted into a multiple-choice question with a prompt-chosen-rejected triple, following the VMR template outlined in Figure \ref{fig:vmr}.
    Rewards are achieved by using rule-based verification functions (adapted from the \texttt{math-verify} package).  
    To avoid trivial data, we filter out MCQuestions whose model accuracy falls outside the range $[0\%, 85\%]$, forming $\mathcal{D}_{\text{MCQ}}$.

    \item \textbf{MCPrompts Dataset}
    is a collection of prompts extracted from the MCQuestions dataset.
    Rewards are assigned by the {URM-LLaMA-3.1-8B} reward model \cite{DBLP:journals/corr/abs-2410-00847}, trained on datasets such as HelpSteer2 \cite{DBLP:conf/iclr/WangBDESZKD25} and Skywork \cite{DBLP:journals/corr/abs-2410-18451}. It is referred to as $\mathcal{D}_{\text{MCP}}$, and $\mathcal{D}_{\text{MCP}}^{\text{sub}}$ is derived from a 5\% subsample of dataset $\mathcal{D}_{\text{MCP}}$.

    \label{section:GeneralPrompts}
    \item \textbf{GeneralPrompts Dataset} contains approximately 20k prompts from diverse sources, including \textit{Awesome-ChatGPT-Prompts}, \textit{Roleplay-Instructions-Dataset}, \textit{Roleplay-Hausa} and \textit{Tulu-3-Sft} \cite{DBLP:journals/corr/abs-2411-15124}. 
    Prompts include question answering, creative writing, instruction following, and role-playing.  Rewards are assigned by the {URM-LLaMA-3.1-8B} reward model \cite{DBLP:journals/corr/abs-2410-00847}, trained on datasets such as HelpSteer2 \cite{DBLP:conf/iclr/WangBDESZKD25} and Skywork \cite{DBLP:journals/corr/abs-2410-18451}. It is labeled as $\mathcal{D}_{\text{GP}}$, and $\mathcal{D}_{\text{GP}}^{\text{sub}}$ is derived from a 5\% subsample of dataset $\mathcal{D}_{\text{GP}}$.
\end{itemize}

\textbf{Baselines.}  
We compare VMR‑RLVR with traditional RLHF under identical training configurations and sample sizes, with the training data and reward mechanism being the sole differing factors.
\begin{itemize}
    \item \textbf{RLHF with $\mathcal{D}_{\text{MCP}}$} optimizes policy model on the prompts from $\mathcal{D}_{\text{MCP}}$.
    \item \textbf{RLHF with $\mathcal{D}_{\text{GP}}$} trains the policy model on the prompts from $\mathcal{D}_{\text{GP}}$.
\end{itemize}

\begin{table*}[t]
\caption{Overall performance on benchmarks comparing baselines with VMR-RLVR.
In the \textit{Model column}, 
\textbf{RLHF w/$\mathcal{D}_{\text{MCP}}$} uses prompts extracted from the MCPrompts dataset and discards chosen and rejected responses;
\textbf{RLHF w/$\mathcal{D}_{\text{GP}}$} trains a policy model on prompts from the GeneralPrompts dataset;
\textbf{VMR-RLVR} fits a policy model on the prompts from $\mathcal{D}_{\text{MCQ}} \cup \mathcal{D}_{\text{MCP}}^{\text{sub}}$ or $\mathcal{D}_{\text{MCQ}} \cup \mathcal{D}_{\text{GP}}^{\text{sub}}$. It uses rule-based verifiers for  $\mathcal{D}_{\text{MCQ}}$ and model-based rewards for $\mathcal{D}_{\text{MCP}}^{\text{sub}}$ or $\mathcal{D}_{\text{GP}}^{\text{sub}}$.
Best values are in \textbf{\underline{bold}}. All results are averages over 4 runs (\textbf{Avg@4}). }
\resizebox{1.0\linewidth}{!}{

\begin{tabular}{
>{\columncolor[HTML]{FFFFFF}}c 
>{\columncolor[HTML]{FFFFFF}}c 
>{\columncolor[HTML]{FFFFFF}}c 
>{\columncolor[HTML]{FFFFFF}}c 
>{\columncolor[HTML]{FFFFFF}}c 
>{\columncolor[HTML]{FFFFFF}}c 
>{\columncolor[HTML]{FFFFFF}}c 
>{\columncolor[HTML]{FFFFFF}}c 
>{\columncolor[HTML]{FFFFFF}}c 
>{\columncolor[HTML]{FFFFFF}}c }
\hline
{\color[HTML]{1C1D1F} \textbf{Model}} & \textbf{Paradigm}     & {\color[HTML]{1C1D1F} \textbf{\begin{tabular}[c]{@{}c@{}}MT\\ Bench\end{tabular}}} & {\color[HTML]{1C1D1F} \textbf{\begin{tabular}[c]{@{}c@{}}Alpaca\\ Eval\\ 2.0\end{tabular}}} & {\color[HTML]{1C1D1F} \textbf{\begin{tabular}[c]{@{}c@{}}Wild\\ Bench\end{tabular}}} & {\color[HTML]{1C1D1F} \textbf{\begin{tabular}[c]{@{}c@{}}Creative\\ Writing\\ V3\end{tabular}}} & {\color[HTML]{1C1D1F} \textbf{\begin{tabular}[c]{@{}c@{}}ArenaHard2.0\\ Creative\\ Writing\end{tabular}}} & {\color[HTML]{1C1D1F} \textbf{IFEval}} & {\color[HTML]{1C1D1F} \textbf{\begin{tabular}[c]{@{}c@{}}Live\\ Bench\\ IFEval\end{tabular}}} & \textbf{Avg}   \\ \hline
DS-R1-Distill-Qwen-32B                             & -                     & 7.65                                                                               & 18.86                                                                                       & 2.19                                                                                 & 57.08                                                                                           & 6.17                                                                                                      & 75.23                                  & 73.83                                                                                         & 34.43          \\ \hline
\multicolumn{10}{l}{\cellcolor[HTML]{FFFFFF}{\color[HTML]{000000} {\ul \textit{\colorbox{purple!10}{Reasoning Model} | DeepSeek-R1-Distill-Qwen-14B}}}}                                                                                                                                                                                                                                                                                                                                                                                                                                                                                                                                                                               \\
DS-R1-Distill-Qwen-14B                             & -                     & 7.32                                                                               & 15.54                                                                                       & 1.98                                                                                 & 55.60                                                                                           & 2.50                                                                                                      & 72.58                                  & 73.10                                                                                         & 32.66          \\

\rowcolor[HTML]{ECF4FF} 
RLHF w/$\mathcal{D}_{\text{MCP}}$                                                                           & RLHF                  & 6.91                                                                               & 21.54                                                                                       & 1.00                                                                                 & 63.84                                                                                           & 3.80                                                                                             & 67.28                                  & 72.61                                                                                         & 33.85 {\scriptsize+0.00}          \\
\rowcolor[HTML]{ECF4FF} 
\textbf{Ours ($\mathcal{D}_{\text{MCQ}} \cup \mathcal{D}_{\text{MCP}}^{\text{sub}}$)} & \textbf{VMR-RLVR} & 7.75                                                                      & 25.14                                                                              & 1.78                                                                                 & 68.52                                                                                  & 3.50                                                                                                      & 69.32                         & 73.21                                                                                & 35.60 {\scriptsize+1.75} \\
\rowcolor[HTML]{FEFDE0} 
RLHF w/$\mathcal{D}_{\text{GP}}$                 & RLHF                  & 7.88                                                                               & 28.78                                                                                       & 1.57                                                                                 & 69.52                                                                                           & 4.82                                                                                                      & 75.37                                  & 74.35                                                                                         & 37.47 {\scriptsize+0.00}          \\
\rowcolor[HTML]{FEFDE0} 
\textbf{Ours ($\mathcal{D}_{\text{MCQ}} \cup \mathcal{D}_{\text{GP}}^{\text{sub}}$)}              & \textbf{VMR-RLVR} & \colorbox{green!10}{\textbf{\underline{8.43}}}                                                                      & \colorbox{green!10}{\textbf{\underline{39.41} }}                                                                             & \colorbox{green!10}{\textbf{\underline{3.28}    }}                                                                    & \colorbox{green!10}{\textbf{\underline{75.65}  }}                                                                                & \colorbox{green!10}{\textbf{\underline{6.22}}}                                                                                             & \colorbox{green!10}{\textbf{\underline{76.71}}}                         & \colorbox{green!10}{\textbf{\underline{75.65}}}                                                                                & \colorbox{green!10}{\textbf{\underline{40.76 {\scriptsize+3.29}}}} \\ \hline
\multicolumn{10}{l}{\cellcolor[HTML]{FFFFFF}{\ul \textit{\colorbox{purple!10}{Reasoning Model} | DeepSeek-R1-Distill-Llama-8B}}}                                                                                                                                                                                                                                                                                                                                                                                                                                                                                                                                                                                                     \\
DS-R1-Distill-Llama-8B                              & -                     & 6.06                                                                               & 11.46                                                                                       & 1.44                                                                                 & 49.70                                                                                           & 1.40                                                                                                      & 61.00                                  & 71.58                                                                                         & 28.95        \\
\rowcolor[HTML]{FEFDE0} 
RLHF w/$\mathcal{D}_{\text{GP}}$                 & RLHF                  & 6.13                                                                               & 11.28                                                                                       & 1.00                                                                                 & 52.48                                                                                           & 1.30                                                                                                      & 66.54                                  & 64.22                                                                                         & 28.99 {\scriptsize+0.00}           \\
\rowcolor[HTML]{FEFDE0} 
\textbf{Ours ($\mathcal{D}_{\text{MCQ}} \cup \mathcal{D}_{\text{GP}}^{\text{sub}}$)}              & \textbf{VMR-RLVR} & \colorbox{green!10}{\textbf{\underline{6.53}    }}                                                                  & \colorbox{green!10}{\textbf{\underline{11.83} }}                                                                             & \colorbox{green!10}{\textbf{\underline{1.69}}}                                                                        & \colorbox{green!10}{\textbf{\underline{53.45}}}                                                                                  & \colorbox{green!10}{\textbf{\underline{1.90}}}                                                                                             & \colorbox{green!10}{\textbf{\underline{70.79}}}                         & \colorbox{green!10}{\textbf{\underline{72.86}}}                                                                                & \colorbox{green!10}{\textbf{\underline{31.29 {\scriptsize+2.30}  }}}          \\ \hline
\multicolumn{10}{l}{\cellcolor[HTML]{FFFFFF}{\ul \textit{\colorbox{blue!10}{Non-Reasoning Model} | Qwen2.5-14B-Instruct}}}                                                                                                                                                                                                                                                                                                                                                                                                                                                                                                                                                                                                          \\
Qwen2.5-14B-Instruct                       & -                     & 8.61                                                                               & 39.22                                                                                       & 5.17                                                                                 & 53.05                                                                                           & 3.70                                                                                                      & 78.56                                  & 59.41                                                                                         & 35.39          \\
\rowcolor[HTML]{FEFDE0} 
RLHF w/$\mathcal{D}_{\text{GP}}$                & RLHF                  & 8.75                                                                               & \colorbox{green!10}{\textbf{\underline{48.30}}}                                                                              & \colorbox{green!10}{\textbf{\underline{5.62}}}                                                                        & \colorbox{green!10}{\textbf{\underline{57.68} }}                                                                                 & 3.81                                                                                                      & 79.67                                  & 61.62                                                                                         & 37.92 {\scriptsize+0.00}         \\
\rowcolor[HTML]{FEFDE0} 
\textbf{Ours ($\mathcal{D}_{\text{MCQ}} \cup \mathcal{D}_{\text{GP}}^{\text{sub}}$)}              & \textbf{VMR-RLVR} & \colorbox{green!10}{\textbf{\underline{8.85}   }}                                                                   & 45.47                                                                                       & 5.39                                                                                 & 57.40                                                                                           & \colorbox{green!10}{\textbf{\underline{6.40}  }}                                                                                           & \colorbox{green!10}{\textbf{\underline{79.85} }}                        & \colorbox{green!10}{\textbf{\underline{65.57}   }}                                                                             & \colorbox{green!10}{\textbf{\underline{38.42{\scriptsize+0.50}}}} \\ \hline
\end{tabular}

}

\label{OverallTable}
\end{table*}

\textbf{Implementation Details.} We utilize the Verl framework \cite{DBLP:conf/eurosys/ShengZYWZZPL025}. In each rollout stage, we generate 16 responses from 512 prompts, maintaining a temperature and top-p value of 1.0, without implementing the dynamic sampling method. Subsequently, we conduct 16 policy updates using these responses. We clip the ratio within the range of 0.8 to 1.24 and set the clip-ratio-c \cite{DBLP:conf/aaai/YeLSSZWYYWGCYZS20} to 10.0 to avert entropy collapse. We calculate the average loss using the sequence-mean-token-mean method. We do not incorporate KL divergence into either the rewards or the final loss calculation. We set the entropy coefficient to 0.0 and set the learning rate to 1e-6. We constrain the maximum prompt length and decode length to 16,384 tokens, with a total length limit of 32,768 tokens. The $\alpha$ is set to 0.1. Each experiment is conducted using 32 NVIDIA H800 GPUs, each equipped with 80GB of memory, and lasts for approximately 12 hours.

\textbf{Evaluation Benchmarks.} We evaluate our models on multiple benchmarks. For general chat domains, we include MTBench \cite{DBLP:conf/nips/ZhengC00WZL0LXZ23}, AlpacaEval2.0 \cite{alpaca_eval} and WildBench \cite{DBLP:conf/iclr/LinDCRPD0025}. Whenever possible, we extract prompts related to open-ended subcategories. For creative writing, we use CreativeWritingV3 Benchmark \cite{creative-writing-bench-v3} and ArenaHard2.0-CreativeWriting \cite{DBLP:conf/icml/LiCFD0ZGS25}. To evaluate the ability to follow instructions in open-ended contexts, we include IFEval \cite{DBLP:journals/corr/abs-2311-07911} and LiveBench-IFEval \cite{DBLP:journals/corr/abs-2406-19314}. 

\textbf{Evaluation Configurations.} Reasoning models run in their thinking mode with the rollout temperature set to 0.6 and top-p set to 0.95. Non-reasoning models run in their non-thinking mode with the rollout temperature set to 0.7 and top-p set to 0.8.
To reduce the evaluation variance, we evaluate the model on each benchmark multiple times and report the final Avg@4 results. For reliable
answer extraction, we adopt the $\textless \text{think}\textgreater \ \texttt{THINK} \ \textless \text{/think}\textgreater \ \texttt{RESPONSE}$ template of DeepSeek-R1 \cite{DBLP:journals/corr/abs-2501-12948} and use the \texttt{RESPONSE} part as the generated answer. The max decoding length for training is 32,768, with minimal truncation observed.




\subsection{Main Results}
Table~\ref{OverallTable} summarizes the primary evaluation results. We benchmark our proposed VMR-RLVR framework against a baseline of traditional RLHF. The evaluation encompasses three model families, categorized into reasoning models (DeepSeek-R1-Distill-Qwen-14B and DeepSeek-R1-Distill-Llama-8B) and a non-reasoning instructing model (Qwen2.5-14B-Instruct), to assess the method's efficacy in enhancing open-ended reasoning.

\textbf{VMR-RLVR demonstrates superior performance across reasoning models, significantly enhancing open-ended reasoning capabilities.} 
Our approach consistently surpasses the traditional RLHF baseline across all reasoning model series. Specifically, on the DeepSeek-R1-Distill-Qwen-14B architecture, VMR-RLVR achieves an average score improvement of 1.75 points over the baseline (35.60 vs. 33.85) under identical prompt settings. This performance gap widens to 3.29 points (40.76 vs. 37.47) when introducing a wider variety of prompts. Similarly, for the DeepSeek-R1-Distill-Llama-8B model, our method increases the average score by 2.30 points (31.29 vs. 28.99). Strikingly, our VMR-RLVR method enables the 14B model outperform the much larger 32B benchmark (Avg 34.43) by a margin of 6.33 points, demonstrating significant parameter efficiency.

\textbf{Improvements are marginal on non-reasoning models.} In the case of Qwen2.5-14B-Instruct, VMR-RLVR yields a modest average increase of only 0.50 points, outperforming the baseline on merely four benchmarks. This distinct disparity in performance gains suggests that intrinsic strong reasoning capabilities are a prerequisite for VMR-RLVR to effectively optimize open-ended performance.

\begin{table*}[t]


\caption{Ablation of added extra preference signals in multiple-choice datasets.
\textbf{RLHF w/$\mathcal{D}_{\text{GP}}$} trains only on prompts from the GeneralPrompts dataset;
\textbf{DPO} uses the extracted prompt-chosen-rejected triplets for DPO training.
\textbf{VMR-RLVR} fits the policy model on the prompts from $\mathcal{D}_{\text{MCQ}} \cup \mathcal{D}_{\text{GP}}^{\text{sub}}$. It uses rule-based verifiers for $\mathcal{D}_{\text{MCQ}}$ and model-based rewards for $\mathcal{D}_{\text{GP}}^{\text{sub}}$.
Rows marked with “+” denote ablations. The best numbers are \textbf{bold}. 
All results are averages over 4 runs (\textbf{Avg@4}). }
\resizebox{1.0\linewidth}{!}{

\begin{tabular}{lccccccccc}
\hline
\rowcolor[HTML]{FFFFFF} 
\multicolumn{1}{c}{\cellcolor[HTML]{FFFFFF}{\color[HTML]{1C1D1F} \textbf{Model}}} & \textbf{Paradigm}     & {\color[HTML]{1C1D1F} \textbf{\begin{tabular}[c]{@{}c@{}}MT\\ Bench\end{tabular}}} & {\color[HTML]{1C1D1F} \textbf{\begin{tabular}[c]{@{}c@{}}Alpaca\\ Eval\\ 2.0\end{tabular}}} & {\color[HTML]{1C1D1F} \textbf{\begin{tabular}[c]{@{}c@{}}Wild\\ Bench\end{tabular}}} & {\color[HTML]{1C1D1F} \textbf{\begin{tabular}[c]{@{}c@{}}Creative\\ Writing\\ V3\end{tabular}}} & {\color[HTML]{1C1D1F} \textbf{\begin{tabular}[c]{@{}c@{}}ArenaHard2.0\\ Creative\\ Writing\end{tabular}}} & {\color[HTML]{1C1D1F} \textbf{IFEval}} & {\color[HTML]{1C1D1F} \textbf{\begin{tabular}[c]{@{}c@{}}Live\\ Bench\\ IFEval\end{tabular}}} & \textbf{Avg}   \\ \hline
\rowcolor[HTML]{FFFFFF} 
DeepSeek-R1-Distill-Qwen-14B                                                      & -                     & 7.32                                                                               & 15.54                                                                                       & 1.98                                                                                 & 55.60                                                                                           & 2.50                                                                                                      & 72.58                                  & 73.10                                                                                         & 32.66          \\
\rowcolor[HTML]{EFEFEF} 
RLHF w/$\mathcal{D}_{\text{GP}}$                                                            & RLHF                  & 7.88                                                                               & 28.78                                                                                       & 1.57                                                                                 & 69.52                                                                                           & 4.82                                                                                                      & 75.37                                  & 74.35                                                                                         & 37.47          \\
\rowcolor[HTML]{FFFFFF} 
DPO w/$\mathcal{D}_{DPO}$                                                                            & DPO                   & 7.36                                                                               & 17.92                                                                                       & 2.07                                                                                 & 58.28                                                                                           & 4.40                                                                                                      & 73.38                                  & 72.98                                                                                         & 33.77          \\
\rowcolor[HTML]{EFEFEF} 
\textbf{Ours ($\mathcal{D}_{\text{MCQ}} \cup \mathcal{D}_{\text{GP}}^{\text{sub}}$)}                                                          & \textbf{VMR-RLVR} & \textbf{8.43}                                                                      & \textbf{39.41}                                                                              & \textbf{3.28}                                                                        & \textbf{75.65}                                                                                  & \textbf{6.22}                                                                                             & \textbf{76.71}                         & \textbf{75.65}                                                                                & \textbf{40.76} \\ \hline
\end{tabular}
}

\label{tab:abation}
\end{table*}

\subsection{Ablation studies}
\label{section:ablation}
To determine whether the performance gains of VMR-RLVR represent a fundamental improvement or merely stem from the inclusion of additional preference annotations, we conducted an ablation study comparing our method against a standard Direct Preference Optimization (DPO) baseline.
\begin{itemize}
    \item \textbf{DPO:} We construct a dataset of <query, chosen, rejected> triples derived from the MCQuestions dataset, named as $\mathcal{D}_{DPO}$. We then employ Direct Preference Optimization~\cite{DBLP:conf/nips/RafailovSMMEF23} to optimize the policy by explicitly maximizing the likelihood margin between the chosen and rejected responses.
\end{itemize}

%

\textbf{Impact of pairwise preference signals.} As detailed in Table~\ref{tab:abation}, the DPO baseline yields a score of 33.77. While this represents a marginal improvement over the base model (32.66), it significantly lags behind both the traditional RLHF trained on $\mathcal{D}_{GP}$ (37.47) and our proposed VMR-RLVR (40.76). This disparity indicates that simply ingesting preference knowledge via DPO is insufficient to replicate the efficacy of our framework.


We attribute the suboptimal performance of DPO to the limitations of off-policy training in this context. Since the chosen and rejected responses are not generated by the model itself, DPO tends to overfit the static data distribution without effectively generalizing to the model's own generation trajectory.

In contrast, VMR-RLVR leverages these preference signals to abstract a robust verification mechanism ("reasoning by contrast"). Rather than merely imitating the token probability distribution of the dataset, VMR-RLVR learns to identify intrinsic patterns that distinguish superior reasoning. Our results suggest that this learned verification capability generalizes effectively across different training regimes, enabling the model to optimize its open-ended reasoning performance far more effectively than direct probability manipulation. For a more granular analysis and case studies, please refer to Section~\ref{sec:analyse}.

\section{Analysis}
\label{sec:analyse}
\subsection{Length Bias and Reasoning Density} 
\begin{figure}
    \centering
    \vspace{-10pt}
    \includegraphics[width=0.99\linewidth]{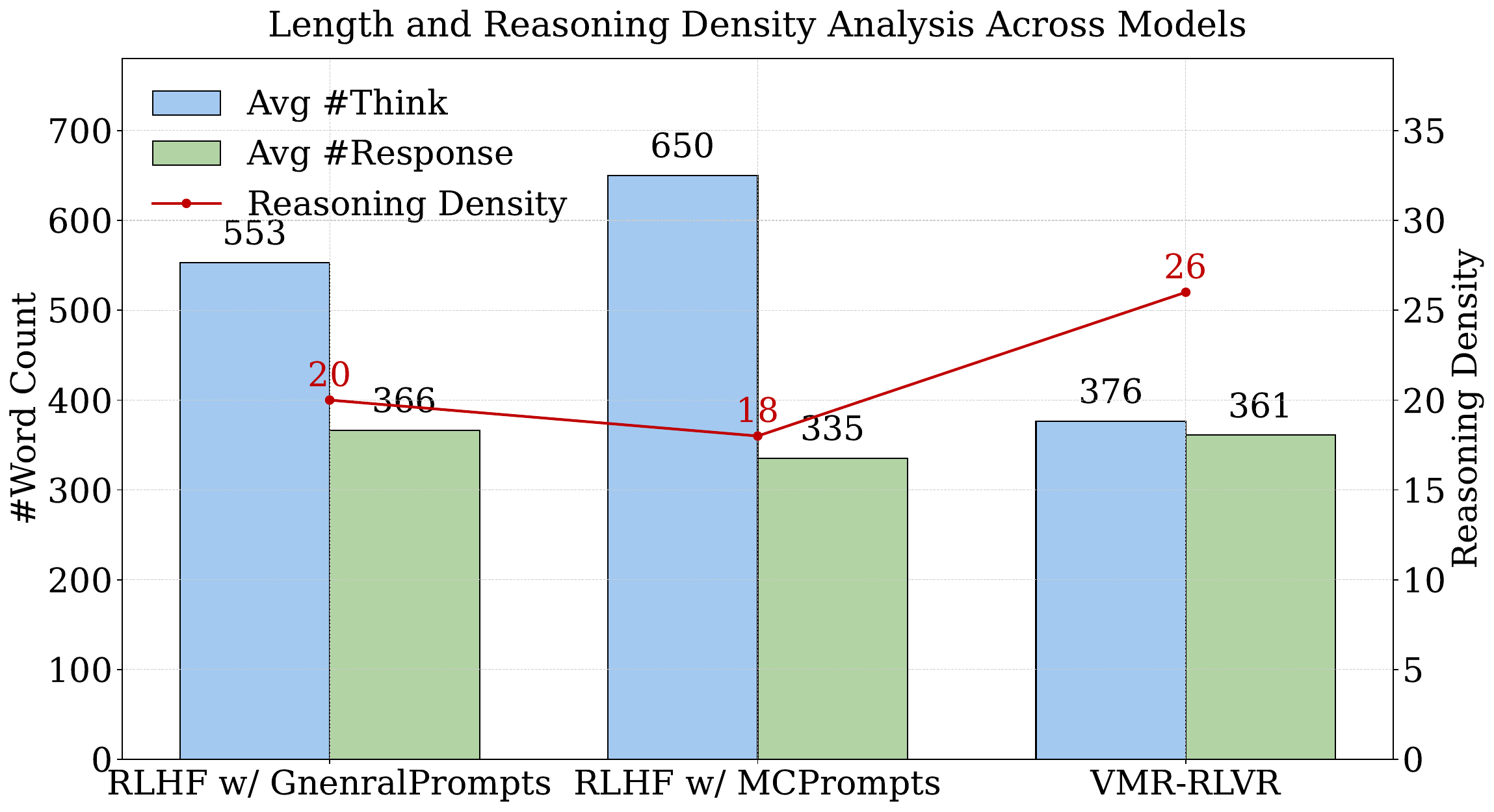}
    \vspace{-3pt}
    \caption{Analysis of length and reasoning density.} \label{DensityAnalysis}             
    \vspace{-5pt}
\end{figure}

Evaluation in open-ended domain tasks depends heavily on the LLM-as-a-judge paradigm. 
However, it is prone to a potential \textit{length bias}, a known phenomenon where evaluators disproportionately favor longer responses regardless of quality~\cite{DBLP:journals/corr/abs-2408-13006}. To verify that the gains of VMR-RLVR are not an artifact of mere verbosity, we analyzed the token distribution of the \texttt{think} and \texttt{response} components across models (Figure~\ref{DensityAnalysis}). 

\textbf{Length Neutrality.} 
We observe that the proposed VMR-RLVR produces responses of comparable length to the RL trained on GeneralPrompts $\mathcal{D}_{GP}$. The absence of significant length inflation suggests that the performance gains are not an \textit{artifact of superficial verbosity}, but rather stem from \textit{enhanced reasoning quality}.

\textbf{Reasoning Density.} 
To further quantify the efficiency of the generated chain-of-thought, we introduce the metric of \textit{Reasoning Density}, defined as the number of distinct reasoning steps per 1,000 words (identified via the zero-shot prompt detailed in Table~\ref{tab:density-prompt}). As illustrated in Figure~\ref{DensityAnalysis}, VMR-RLVR exhibits a significantly higher reasoning density compared to both the RLHF w/$\mathcal{D}_{GP}$ and RLHF w/$\mathcal{D}_{MCQ}$. This demonstrates that within a similar token budget, VMR-RLVR encapsulates more structured reasoning steps, reflecting a more deliberate and information-dense problem-solving process.




\subsection{Case study} 
We conducted a qualitative analysis of the \texttt{THINK} process to elucidate the mechanisms underlying the observed improvements. Compared to the RL w/$\mathcal{D}_{GP}$, VMR-RLVR enhances reasoning in three key dimensions: first, it \textbf{mitigates redundancy} by eliminating rote repetition of context and discarding generic, context-agnostic statements; second, it generates \textbf{highly prompt-specific analysis} characterized by greater depth and substance; and third, it more \textbf{accurately discerns user intent}, particularly regarding implicit or derived nuances. Table~\ref{tab:case-study} presents detailed examples illustrating these improvements.

\subsection{UMAP Visualization} 
To investigate the latent semantic structure of the generated reasoning, we visualize the model outputs in a low-dimensional space using UMAP. We randomly sampled 30 queries from \textit{ArenaHard2.0-CreativeWriting} and \textit{CreativeWriting-V3} respectively and generated five reasoning trajectories per query. These reasoning trajectories are encoded by prevailing \texttt{all-MiniLM-L6-v2}\footnote{\url{https://huggingface.co/sentence-transformers/all-MiniLM-L6-v2}}.

\begin{figure}
    \centering
    \includegraphics[width=0.99\linewidth]
    {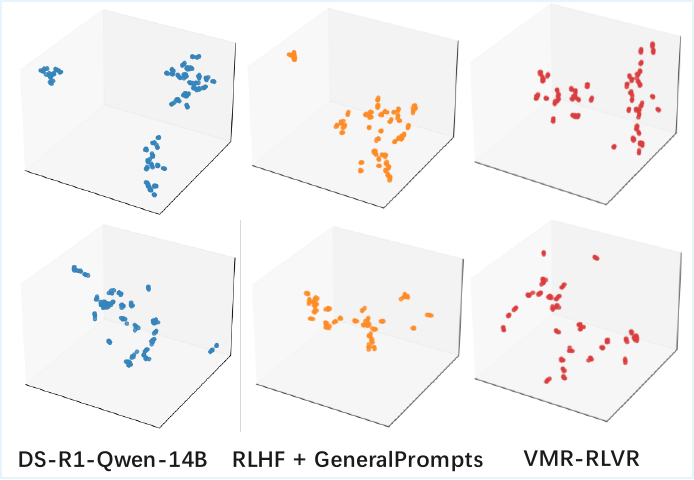}
    \caption{UMAP visualization of embedding distributions. The first row shows results on ArenaHard2.0-CreativeWriting, while the second row shows results on CreativeWriting-V3.}
    \label{fig:umap}
    \vspace{-10pt}
\end{figure}

As illustrated in Figure~\ref{fig:umap}, the baseline models exhibit a highly fragmented topology, characterized by numerous isolated, small-scale clusters. This suggests that the baselines operate in localized reasoning modes, with limited semantic connectivity across different queries. In contrast, our model forms fewer, denser, and more spatially proximal clusters. This distribution indicates that the model has learned to generalize shared reasoning patterns across diverse queries, rather than overfitting to specific prompt instances. Furthermore, the internal dispersion within these larger clusters suggests that while the model converges on robust high-level strategies, it retains sufficient variance to address specific query nuances.

Crucially, this finding corroborates the density analysis in Figure~\ref{DensityAnalysis}: although our reasoning traces are more concise, they occupy a more unified semantic manifold, demonstrating that the method achieves efficiency without sacrificing semantic coverage.

\section{Related Work}
\textbf{Reinforcement Learning with Verifiable Rewards.}
RLVR refers to reinforcement learning methods where the reward is directly computed by task-specific verifiers that check the correctness of model outputs. In mathematical reasoning, the most common design is answer matching, where a binary reward is assigned depending on whether the predicted answer matches the reference solution \cite{DBLP:journals/corr/abs-2501-12599,DBLP:journals/corr/abs-2501-12948,DBLP:journals/corr/abs-2410-15115,DBLP:journals/corr/abs-2411-15124,DBLP:journals/corr/abs-2503-18892,DBLP:journals/corr/abs-2503-10460,DBLP:journals/corr/abs-2503-17287}. Similarly, in code generation tasks, program execution or unit testing is used to automatically verify correctness \cite{luo2025deepcoder,he2025skyworkopenreasoner1,DBLP:journals/corr/abs-2502-01456,DBLP:journals/corr/abs-2508-05170}. These designs eliminate the need for learned reward models and instead rely on deterministic evaluation, which has been shown to greatly stabilize training. Despite these advantages, RLVR is inherently limited to domains where such verifiers exist, restricting its applicability beyond STEM-oriented problems.

\textbf{Reward Models for Open-Ended Tasks.}
In the absence of explicit verifiers, reward models (RMs) trained from human preference annotations have become the dominant approach for aligning LLMs with open-ended tasks, forming the foundation of Reinforcement Learning with Human Feedback (RLHF) \cite{DBLP:conf/nips/Ouyang0JAWMZASR22,DBLP:journals/corr/abs-2204-05862}. While this paradigm has enabled notable progress in instruction following, summarization, and dialogue, it also introduces fundamental challenges. RMs \cite{liu2025skywork,DBLP:journals/corr/abs-2503-05244,DBLP:conf/naacl/LambertPMMLCDKZCSH25,DBLP:journals/corr/abs-2505-10320} require large-scale annotated datasets, are computationally expensive to train, and often encode annotator biases or spurious correlations. Moreover, unlike rule-based verifiers, RMs provide preference-based rather than verifiable feedback, which can be noisy and misaligned with true task quality. These limitations highlight the inherent trade-off of RM–based supervision: it offers scalability in open-ended domains but lacks the reliability of verifiability.

\textbf{Reasoning in Open-Ended Tasks.}
Enhancing the reasoning ability of LLMs has been shown to benefit both reasoning-intensive domains and seemingly non-reasoning tasks \cite{DBLP:journals/corr/abs-2501-12948,DBLP:journals/corr/abs-2507-00432,DBLP:journals/corr/abs-2509-20357}. 
Recent efforts have also attempted to broaden the scope of reasoning beyond core STEM problems to fields such as economics, chemistry, and physics \cite{DBLP:journals/corr/abs-2506-18254,DBLP:journals/corr/abs-2505-14652,DBLP:journals/corr/abs-2505-21493}. 
Specifically, \cite{DBLP:journals/corr/abs-2506-18254} and \cite{DBLP:journals/corr/abs-2505-21493} replace explicit verifiers with probabilistic reward estimation, enabling reinforcement signals without symbolic checkers, while \cite{DBLP:journals/corr/abs-2505-14652} constructs a general-purpose reward model by aggregating diverse datasets with verifiable answers. 
Although these approaches broaden the applicability of reasoning, they remain confined to domains where standard answers exist and correctness can still be objectively verified. Truly open-ended tasks, where outputs are inherently diverse and lack unambiguous evaluation criteria, remain largely underexplored.
Our work addresses this gap by introducing a novel VMR-based training strategy that restructures open-ended supervision into verifiable multiple-choice formats, thereby preserving the advantages of RLVR while extending its applicability to tasks without standard answers.

\section{Conclusion}
In this work, we propose Verifiable Multiple-Choice Reformulation for Reinforcement Learning from Verifiable Rewards (VMR-RLVR), a simple yet effective framework that restructures open-ended domain data into verifiable multiple-choice formats, enabling RLVR to operate beyond strictly verifiable STEM domains. 
Extensive experiments demonstrate that our method not only improves task performance but also enhances the reasoning trajectories. In the future, we intend to investigate alternative strategies to alleviate the collapse of response lengths typically associated with training exclusively on multiple-choice datasets.

\section*{Impact Statement}

This paper introduces Verifiable Multiple-Choice Reformulation (VMR), a novel training strategy designed to enhance the reasoning capabilities of large language models (LLMs) in open-ended tasks where explicit ground-truth solutions are unavailable. While our approach shows promise in improving LLM performance across various benchmarks, it also raises several ethical and societal considerations that must be acknowledged.

From an ethical standpoint, it is crucial to consider the potential biases that may arise from the reformulation process. The choice of options in multiple-choice formats can introduce biases based on cultural, social, or personal perspectives, potentially disadvantaging certain groups. Ensuring fairness and inclusivity in the design of these options is essential to prevent unintended discrimination.

Lastly, the dissemination of our code upon acceptance aims to encourage transparency and reproducibility, but it also necessitates responsible usage by the community. Researchers and developers utilizing our approach should be mindful of the broader impacts of deploying VMR-trained models, especially in sensitive applications where the consequences of misinterpretation or bias can be significant.

In summary, while VMR offers a promising advancement in the field of machine learning, it is imperative to remain vigilant regarding its ethical implications and societal consequences, ensuring its application aligns with values of fairness, inclusivity, and the promotion of diverse thought.

\bibliography{example_paper}
\bibliographystyle{icml2026}

\newpage
\appendix

\onecolumn

\section{Verifiable Multiple-Choice Reformulation (VMR) template}

\begin{transcriptbox}
\texttt{You are an expert evaluator. Given a query, please evaluate which of the two responses is better. If the first response is better, return}\verb|\boxed{A}|\texttt{. If the second response is better, return} \verb|\boxed{B}|.

\vspace{\baselineskip}

\texttt{**Query**}

You serve as a modern history consultant who navigates conversations \ldots 

\vspace{\baselineskip}

\texttt{**Response A**}

\texttt{[Response A Start]}

The claim that the events of September 11, 2001, were an inside job and \ldots 

\texttt{[Response A End]}

\vspace{\baselineskip}

\texttt{**Response B**}

\texttt{[Response B Start]}

The tragic events of September 11, 2001, when terrorist attacks were carried out using  \ldots  

\texttt{[Response B End]}

\vspace{\baselineskip}

\texttt{**Output requirement**}

\texttt{Please put your final answer within} \verb|\boxed{answer}|. \texttt{If the first response is better, return} \verb|\boxed{A}|. \texttt{If the second response is better, return} \verb|\boxed{B}|.

\end{transcriptbox}
\captionsetup{font=normalsize}
\captionof{table}{Verifiable Multiple-Choice Reformulation (VMR) template.}
\label{tab:template}

\section{LLM Zero-Shot Prompt For Reasoning Density.}

\begin{transcriptbox}
    
\texttt{Extract and format reasoning points from a given reasoning process as follows: }

\texttt{Your evaluation should: }

\texttt{1. Identify distinct reasoning steps. }

\texttt{2. Extract each step as a string. }

\texttt{3. Return these as a JSON array. }

\texttt{4. Return } \verb|\boxed{n}| \texttt{, where n is the count of reasoning points. }

\texttt{The given reasoning process:}

 Okey, Let us \ldots
 
\end{transcriptbox}
\captionsetup{font=normalsize}
\captionof{table}{LLM zero-shot prompt for reasoning density.}
\label{tab:density-prompt}

\section{Additional Ablation Study}

\begin{transcriptbox}

\verb!<|begin\_of\_sentence|><|User|>!

\vspace{\baselineskip}

You are an expert evaluator. Given a question, please evaluate which of the 2 answers is better. If the first answer is better, return \boxed{1}; if the second answer is better, return \boxed{2}.

\vspace{\baselineskip}

\texttt{Question}

\vspace{\baselineskip}

Write a response that appropriately completes the request.

\vspace{\baselineskip}

Describe how to prepare a cup of herb tea.

\vspace{\baselineskip}

\vspace{\baselineskip}

\texttt{Answer A}
\vspace{\baselineskip}

\texttt{[Answer A Start]}
\vspace{\baselineskip}

To prepare a cup of herb tea, start by boiling one cup of water in a small saucepan. Once the water has reached a boil, remove it from the heat and add one teaspoon of the desired herbs. Let the mixture steep for 10 minutes, then strain out the herbs using a fine-mesh sieve or coffee filter. Once the tea has cooled, give it a stir and enjoy!
\vspace{\baselineskip}

\texttt{[Answer A End]}
\vspace{\baselineskip}

\texttt{Answer B}
\vspace{\baselineskip}

\texttt{[Answer B Start]}
\vspace{\baselineskip}

To make a cup of herb tea, start by adding 1 teaspoon of dried herbs to a teapot or infuser. Fill the teapot with boiled water and stir. Let the herbs steep for 5-10 minutes, depending on your desired strength, and then carefully remove the herb teas from the teapot or infuser. Add 1 teaspoon of honey or natural sweetener, if desired.
\vspace{\baselineskip}

\texttt{[Answer B End]}
\vspace{\baselineskip}

—
\vspace{\baselineskip}

\texttt{Output requirement:}
\vspace{\baselineskip}

**: The output format should be \boxed{answer number}. If the first answer is better, return \boxed{A}; if the second answer is better, return \boxed{B}.
\vspace{\baselineskip}

 Please reason step by step, and put your final answer within \verb|\\boxed{}|.
 
 \vspace{\baselineskip}
 
\verb!<|Assistant|>!

\vspace{\baselineskip}

\texttt{<think>}

\vspace{\baselineskip}

Okay, I need to evaluate which answer is better for the question "Describe how to prepare a cup of herb tea." The user wants me to return either \boxed{1} or \boxed{2} based on which answer is better.
\vspace{\baselineskip}

First, I'll read both answers carefully.
\vspace{\baselineskip}

Answer A:
\vspace{\baselineskip}

They start by boiling one cup of water in a saucepan. Once boiled, remove from heat and add one teaspoon of herbs. Steep for 10 minutes, strain with a sieve or coffee filter. After cooling, stir and enjoy.
\vspace{\baselineskip}

Answer B:
\vspace{\baselineskip}

They add 1 teaspoon of dried herbs to a teapot or infuser, pour boiled water, stir, steep for 5-10 minutes, remove herbs, and optionally add honey.
\vspace{\baselineskip}

Now, I'll compare the two.
\vspace{\baselineskip}

Answer A uses a saucepan and a sieve. It specifies cooling before stirring. Answer B uses a teapot/infuser and allows for adjusting steeping time. It also suggests adding a sweetener.
\vspace{\baselineskip}

Looking at clarity, both are clear. Answer A is straightforward with specific tools. Answer B is also clear and mentions customization with steeping time and sweetener.
\vspace{\baselineskip}

Practicality: Answer A uses a saucepan, which is common, but some might prefer a teapot. Answer B uses a teapot or infuser, which is standard for tea. Answer A strains with a sieve, which is effective, while Answer B uses an infuser, which keeps herbs separate.
\vspace{\baselineskip}

The key difference is that Answer A specifies cooling before stirring, which might not be necessary. Answer B allows for adjusting steeping time, which is more flexible. Also, Answer B adds a sweetener, which is a personal choice.
\vspace{\baselineskip}

I think both are good, but Answer B might be more flexible and offers customization. However, Answer A is more specific and avoids extra steps like adding sweeteners, which might be better for a neutral preparation.
\vspace{\baselineskip}

Wait, the question is about preparing a cup, not about sweetening. So Answer A might be more neutral and focused on the herbal preparation without adding extras. That could be seen as better because it sticks to the preparation without distractions.
\vspace{\baselineskip}

On the other hand, Answer B is more versatile with steeping time and sweeteners, but the instruction was just to prepare a cup, not to personalize it.
\vspace{\baselineskip}

I'm a bit torn, but I think Answer A is more straightforward and doesn't complicate the process with optional steps. It's more about the herbs and water, which might be more aligned with the question.

\vspace{\baselineskip}

\texttt{</think>}

\vspace{\baselineskip}

The first answer is more straightforward and focused solely on preparing the herb tea without additional steps like adding sweeteners. It uses clear, specific methods and tools, making it better suited for the given instruction.

\vspace{\baselineskip}

\verb!\boxed{B}!

\vspace{\baselineskip}

\verb!<|end\_of\_sentence|>!


\end{transcriptbox}
\captionsetup{font=normalsize}
\captionof{table}{An example of pure VMR-RLVR with multiple-choice questions. The LLM is tasked to select the better response between the given two responses.}
\label{tab:PureRLVRExample}

\section{Case Study}

\begin{transcriptbox}

\textbf{\texttt{**Prompt**}}

\texttt{Write a poem about how I am missing my classes. The poem must have 4 sections marked with SECTION X. Finish the poem with this exact phrase: “Can I get my money back for the classes I missed?}

\vspace{\baselineskip}

\textbf{\texttt{**THINK (RL + GeneralPrompts)**}}

\texttt{<think>Alright, so the user wants...}

\texttt{...}

\texttt{The ending question is important, so I’ll ensure it’s impactful. I’ll avoid being too technical, keeping the language simple and heartfelt.}

\colorbox{red!50}{\texttt{Bad: Non-informative}}

\texttt{...}

\texttt{Maybe the first section can talk about the physical absence, the second about the missed learning opportunities, the third about the emotional impact, and the fourth about the financial aspect, ending with the specified line.}

\colorbox{red!50}{\texttt{Bad: Too short}}

\texttt{...}

\texttt{Maybe the first section sets the scene of absence, the second talks about the missed content, the third about the emotional toll, and the fourth about the financial concern.}

\colorbox{red!50}{\texttt{Bad: Repetition of previous paragraph}}

\colorbox{red!50}{\texttt{Bad: Financial concern}}

\colorbox{red!50}{\texttt{Bad: Failed to get implicit intent}}

\vspace{\baselineskip}

\textbf{\texttt{**THINK (VMR-RLVR)**}}

\texttt{<think>Alright, so the user wants...}

\texttt{...}

\texttt{Maybe the first section can talk about the initial feeling of absence and the quiet of an empty classroom. That sets a somber tone.the second section, I can focus on the personal impact—how each missed class affects their growth and the knowledge they’re missing out on. This makes it more personal and relatable. The third section could delve into the emotional weight, like the guilt and the ticking clock of time passing. It adds depth to their feelings of loss. Finally, the fourth section should lead up to the closing line. I need to build towards that question, showing the desire to make things right or recover what’s lost.}

\colorbox{green!50}{\texttt{Good: Prompt-specific}}

\colorbox{green!50}{\texttt{Good: Concise}}

\colorbox{green!50}{\texttt{Good: Not Financial but regret and lost}}

\colorbox{green!50}{\texttt{Good: User Implicit intent }}

\texttt{...}

\texttt{We also need to ensure the rhyme and rhythm are consistent, maybe using a simple ABAB scheme to keep it flowing smoothly. Each section should have four lines, maintaining structure while conveying the emotions effectively.the user’s request, they might be looking for not just a poem, but a way to express their feelings and maybe find some closure. So, the poem should resonate emotionally and end with a powerful, direct question that encapsulates their frustration and hope for a solution}

\colorbox{green!50}{\texttt{Good: Prompt-specific}}

\colorbox{green!50}{\texttt{Good: Detailed and informative}}

\end{transcriptbox}
\captionsetup{font=normalsize}
\captionof{table}{Examples on how VMR-RLVR enhances reasoning.}
\label{tab:case-study}

\twocolumn

\section{Potential Risks}
Bias amplification: Preference datasets and judge models reflect demographic and cultural norms that may not be universal. VMR can propagate and amplify these biases, leading to systematically favored styles, viewpoints, or dialects.
Misuse in targeted persuasion: Improved reasoning under verifiable objectives might be repurposed to craft persuasive, manipulative messaging, particularly when options are framed to guide user choices.

\section{Use or Create Scientific Artifacts}

In this work, we both use existing scientific artifacts.

\subsection{Cite Creators Of Artifacts}

We cite the creators of all datasets, models, and software we use. For each dataset, we reference its canonical paper and, where available, its datasheet/data card. For each pretrained model or API, we cite the model card or provider documentation and the associated paper. For software libraries and evaluation toolkits, we cite the recommended software citations and include exact versions. Full citations appear in the References; we also summarize sources inline where relevant (e.g., in the Data and Methods sections).

\subsection{Discuss The License For Artifacts}
We document the license or terms of use for every artifact we use and release, and we comply with those terms.
- Datasets used:
  - Dataset: licensed under CC BY 4.0 / CC BY-NC 4.0 / custom TOS.
- Pretrained models/APIs:
  - [Model/Checkpoint]: licensed under Apache-2.0 / MIT / custom model license.

\subsection{Artifact Use Consistent With Intended Use}
Our use of existing artifacts is consistent with their intended use and access conditions.
- We use all third-party datasets and models strictly for research purposes, in line with licenses that restrict usage to research/non-commercial contexts.
- Where datasets specify disallowed uses (e.g., biometric identification, surveillance, discrimination), we do not pursue those use cases.
- We do not attempt to re-identify individuals, circumvent access controls, or merge datasets in ways that would violate original consent or terms.
- For artifacts accessed under “research-only” or “no-redistribution” conditions, we do not redistribute restricted content; we share only derived statistics, features, or scripts.
- For API-based models, we follow provider usage policies, including content restrictions and rate limits, and we do not use model outputs to train commercial systems if prohibited.

For artifacts we release, we specify intended use and access conditions to remain compatible with any upstream restrictions:
- Intended use: [e.g., benchmarking and academic research on {task/domain}].
- Out-of-scope uses: [e.g., high-stakes decision making, surveillance, medical or legal advice, content generation without human oversight].
- Access conditions: [e.g., research-only license, acceptance of EULA, acknowledgement of limitations and safety considerations].

\subsection{Data Contains Personally Identifying Info Or Offensive Content}
For all data we curate or release, we assess PII and potentially offensive content and apply protective measures.
- PII screening and de-identification:
  - Automated detection: regular expressions and rule-based detectors for emails, phone numbers, postal addresses, usernames/handles, and IDs; named-entity recognition to flag person names and locations.
  - Manual review: stratified spot checks of a random sample.
  - Remediation: removal or masking of direct identifiers; replacement of user handles and names with consistent anonymized IDs; coarse-graining of quasi-identifiers (e.g., binning timestamps to date or week).
  - Audit and logging: we track the number and types of redactions and retain only the minimally necessary metadata.
  - Re-identification safeguards: we do not publish raw free text from sources that prohibit redistribution or that may contain PII; we provide derived artifacts when needed.
- Offensive/toxic content handling:
  - Automated filtering: toxicity and hate-speech classifiers [e.g., specify tool/version if applicable] with conservative thresholds; blocklists/allowlists for known slurs and sensitive terms.
  - Human oversight: manual review of borderline cases to reduce over-removal or inadvertent bias; clear annotation guidelines and escalation procedures for annotator safety.
  - Purpose and warning: where retaining harmful content is necessary for research (e.g., safety or toxicity detection), we minimize exposure, clearly label such content, and include content warnings in documentation.
- Data retention and access:
  - We store raw data on secure, access-controlled systems with least-privilege permissions and retain it only as long as necessary for replication.
  - Released datasets exclude PII and adhere to source licenses; access may be gated by a data-use agreement when required.

All of the above procedures are documented in the dataset/model cards accompanying our released artifacts, including known limitations, potential biases, and appropriate uses.

\section{Human Subjects Including Annotators}

\subsection{Instructions Given To Participants}
\label{section-instrution-chinese-human-eval}

1. Evaluation Criteria: When evaluating question-and-answer topics, more emphasis should be placed on the structured nature of the answer and its logical sequence. Answers with a clear structural advantage and logical clarity should be given higher scores.  

2. Supporting Information: Acknowledge the importance of supporting information in QA and decide to properly consider this factor when scoring.  

3. "The More, The Better" Principle: When evaluating QA topics, adhere to the "more is better" principle, but the content of the answer must provide informational value while avoiding redundancy and irrelevant correctness.  

4. Clear Ranking Points: Define three clear ranking points as critical standards for evaluating answer quality:

   1) Increase in title hierarchy  
   2) Structural adjustments and quality enhancements  
   3) Significant increase in dimensional information  
   If there are noticeable improvements in structural adjustments, title hierarchy, Markdown formatting, or additional data without factual errors, the same ranking can be assigned.  

5. A single error in the core fact of the QA will result in a score of zero (considering all constraints, the queried fact itself is defined as the core fact).  

For example: When asking, "What generation ruler of the Qin state was Duke Mu of Qin?" the core fact is "what generation."  \\
If the AI provides the following response: "Duke Mu of Qin was the 15th ruler of Qin. According to Baidu Baike, from the time when Feizi was enfeoffed as a vassal until King Zheng unified the six states to establish the Qin Dynasty, there were a total of 31 rulers of Qin. Duke Mu of Qin reigned from 659 BC to 621 BC and was not only significant in the history of the Qin state but also an influential ruler," the score would be zero because the core fact is incorrect.

\subsection{Recruitment And Payment}
Some new human annotators or participants were recruited for this study. They are outsourced employees and their salaries follow market pricing.

\subsection{Data Consent}
Data sources: We used publicly available datasets released for research under explicit licenses or terms of use (e.g., open-source preference corpora and benchmark prompts).
Consent status: We did not collect new human data. Consent for included human-contributed data is governed by the original dataset agreements. We honor those terms by not redistributing restricted content and by citing sources.
Use explanation: Our reformulation (VMR) converts existing preference pairs or ranked outputs into multiple-choice items. This transformation does not add new personal information and is limited to research purposes consistent with the source licenses.
Additional protections: We applied PII scrubbing and safety filtering as described above, and we will release only the transformation code and indices, not raw sensitive data. Users must obtain original datasets from their sources and accept their licenses before running our pipeline.

\subsection{Ethics Review Board Approval}
We follow the ethical guidelines set by the Code of Ethics. Our research uses publicly accessible datasets that are properly licensed. We have adhered to all usage terms and ensured that no personal or sensitive data was gathered or analyzed. All experiments were carried out using institutional resources, in full compliance with legal and data management regulations. As part of the submission process, we will disclose our funding sources and any potential conflicts of interest.

\subsection{Characteristics Of Annotators}
All evaluators are Chinese.

\section{Ai Assistants In Research Or Writing}
\subsection{Information About Use Of Ai Assistants}
No ideas, methods, analyses, or conclusions were generated or influenced by the LLM. All research design, experiments, analysis, and interpretations are solely the work of the listed authors.

In Section 3, when analyzing reasoning density, we utilize a large language model to determine the number of reasoning steps within the \texttt{think} component.

\section{Ethics statement}
We follow the ethical guidelines set by the Code of Ethics. Our research uses publicly accessible datasets that are properly licensed. We have adhered to all usage terms and ensured that no personal or sensitive data was gathered or analyzed. All experiments were carried out using institutional resources, in full compliance with legal and data management regulations. As part of the submission process, we will disclose our funding sources and any potential conflicts of interest.

\section{Reproducibility statement}
In Section 3, we provide all the details needed to replicate our experiments. This includes information on training data, methods used, the training framework, hyperparameters, experimental setups, evaluation techniques, and decoding settings. We have constructed our implementation using publicly accessible frameworks and have thoroughly documented every experimental configuration. This allows the research community to verify and expand upon our work. To ensure reliable statistical outcomes, all reported results are averaged across multiple trials.

\end{document}

%% file: math_commands.tex
\usepackage{amsmath,amsfonts,bm}

\def\eqref#1{equation~\ref{#1}}

\def\1{\bm{1}}

\DeclareMathAlphabet{\mathsfit}{\encodingdefault}{\sfdefault}{m}{sl}
\SetMathAlphabet{\mathsfit}{bold}{\encodingdefault}{\sfdefault}{bx}{n}

